\documentclass{article}
\pdfoutput=1

\PassOptionsToPackage{numbers, compress}{natbib}
\usepackage[preprint]{nips_2018}

\usepackage[utf8]{inputenc} % allow utf-8 input
\usepackage[T1]{fontenc}    % use 8-bit T1 fonts
\usepackage[colorlinks=true]{hyperref}       % hyperlinks
\usepackage{url}            % simple URL typesetting
\usepackage{booktabs}       % professional-quality tables
\usepackage{amsfonts}       % blackboard math symbols
\usepackage{nicefrac}       % compact symbols for 1/2, etc.
\usepackage{microtype}      % microtypography
\usepackage{expl3}
\usepackage{graphicx}
\usepackage{wrapfig}
\usepackage{multirow}
\usepackage{color}
\usepackage[colorinlistoftodos]{todonotes}
\usepackage{algorithm}
\usepackage{algorithmic}
\setcitestyle{square}

\ExplSyntaxOn
\newcommand\latinabbrev[1]{
  \peek_meaning:NTF . {% Same as \@ifnextchar
    #1\@}%
  { \peek_catcode:NTF a {% Check whether next char has same catcode as \'a, i.e., is a letter
      #1.\@ }%
    {#1.\@}}}
\ExplSyntaxOff

\newcommand\blfootnote[1]{%
  \begingroup
  \renewcommand\thefootnote{}\footnote{#1}%
  \addtocounter{footnote}{-1}%
  \endgroup
}

\def\eg{\latinabbrev{\emph{e.g}}}
\def\ie{\latinabbrev{\emph{i.e}}}
\def\etal{\latinabbrev{\emph{et al}}}
\def\etc{\latinabbrev{\emph{etc}}}

\title{STS Classification with Dual-stream CNN}

\author{
  $\textrm{Shuchen Weng}^{1*} \quad \textrm{Wenbo Li}^{2*} \quad \textrm{Yi Zhang}^{1} \quad \textrm{Siwei Lyu}^{2}$\\
  $\textrm{}^{1}$Tianjin University$\quad\textrm{}^{2}$University at Albany, SUNY\\
}

\begin{document}

\maketitle

\blfootnote{* indicates equal contributions.}
\begin{abstract}
The structured time series (STS) classification problem requires the modeling of interweaved spatiotemporal dependency. most previous STS classification methods model the spatial and temporal dependencies independently. Due to the complexity of the STS data, we argue that a desirable STS classification method should be a holistic framework that can be made as adaptive and flexible as possible. This motivates us to design a deep neural network with such merits. Inspired by the dual-stream hypothesis in neural science, we propose a novel dual-stream framework for modeling the interweaved spatiotemporal dependency, and develop a convolutional neural network within this framework that aims to achieve high adaptability and flexibility in STS configurations from various diagonals, \ie, sequential order, dependency range and features. The proposed architecture is highly modularized and scalable, making it easy to be adapted to specific tasks. The effectiveness of our model is demonstrated through experiments on synthetic data as well as benchmark datasets for skeleton based activity recognition.
\end{abstract}

\section{Introduction}
\label{sec:intro}
% 1. explain the importance of time series classification,
%   define time series, and introduce structured time series
Time series are an important and ubiquitous source of data that correspond to sequences of observations ordered in time \cite{SantosK16}. Time series data are produced from a wide range of natural phenomena and human activities such as weather readings, stock prices, physiological signals and human motions. The observations in many important practical types of time series data are of high dimensional nature. For instance, the financial time series usually include stocks of various companies, and in computer vision, human actions can be represented as a concatenated vector of 3D locations of all joints of a human skeleton. In addition, the components in each observation usually also have strong statistical dependencies: the prices of stocks of different companies are not independent from each other, and so is the case for locations of joints of a human body. Multivariate time series with explicit statistical dependencies among different components are known as structured time series (STS) \cite{GongMZ14}.

Time series classification is the problem of categorizing different time series into pre-defined classes, and it has applications in many areas and fields such as finance, industry, security and healthcare.  One important aspect of STS classification is that the statistical dependencies in the spatial and temporal domains are usually intertwined. However, most previous STS classification methods model the spatial and temporal dependencies independently. Furthermore, these methods typically focus on improving individual steps in a STS classification pipeline, \eg, explicitly modeling spatiotemporal information \cite{LiuWHDK17,SongLXZL17,WangW17,WengWY17}, increasing sequential orders \cite{DuWW15,LiuSXW16,LiuWHDK17}, learning discriminative features \cite{DuWW15,HuangWPG17,KeBASB17,LiuSXW16,LiuWHDK17,RahmaniB17,SongLXZL17,WengWY17,ZhangLXZXZ17}, and adjusting sequential dependency ranges \cite{KeBASB17,LeeKKL17}. Due to the complexity of the STS data, we argue that a desirable STS classification method should be a holistic framework that can be made as adaptive and flexible as possible. This motivates us to design a deep neural network with such merits.

We take inspiration from the the {\em dual-stream neural processing hypothesis} of human visual neural system \cite{goodale1992separate,norman2002two}, which states that there exist two distinct neural processing streams that process the inter-dependent spatiotemporal visual signals. Specifically, the {\em ventral} stream (or the structural stream) specializes for structures that are relatively invariant in time, and requires long-term memory for representations. The {\em dorsal} stream (or the temporal stream) emphasizes on short-term temporal variations, which only requires the short-term memory for representations.

\begin{figure*}
\centering
\includegraphics[width=1\textwidth]{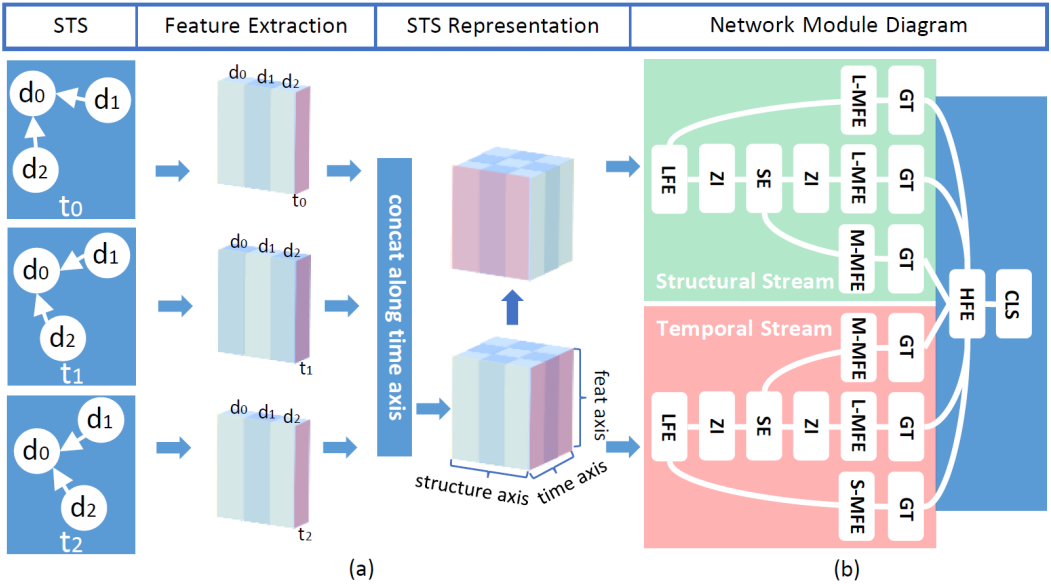}
~\vspace{-1em}
\caption{\small \em (a) \emph{\textbf{The STS representation}} is a tensor with three dimensions, which is formed in two steps. We represent each dimension at each time step as a feature vector, and concatenate them according to the dimension and time order. (b) \emph{\textbf{Module diagram}}. The red and green blocks represent the temporal and structural stream, respectively. L/M/HFE mean the low/medium/high-level feature extractor, and the prefixes of MFE, S/M/L, mean the short/medium/long range. ZI means the zoom-in module, and SE means the encoder shared by M-MFE and L-MFE. GT and CLS represent the gating and classification module, respectively. See texts for details.}
\label{fig:module_diagram}
~\vspace{-1.5em}
\end{figure*}

Following a similar methodology, we propose a dual-stream convolutional neural network for STS classification (as shown in Figure~\ref{fig:module_diagram}). The dual-stream neural network explicitly captures the spatiotemporal dependency in STS data, and the two streams of the network focus on structural and temporal dependencies of different scales. This endows our model higher adaptability and flexibility in modeling the spatiotemporal relationships of input elements, compared to the single-stream \cite{LiuWHDK17,SongLXZL17,WangW17,WengWY17} based methods.  Specifically, a STS instance is first transformed into a 3-rank tensor, with the three dimensions correspond to the time steps, spatial structure and descriptive features of each sample, respectively.  The two streams are implemented as convolutional neural network so that it is not limited to forward or backward directions as typical sequential models (e.g., RNNs) \cite{GehringAGYD17}. The dual-stream CNN model is based on a set of {\em dual-stream convolution kernels}, each formed as a tensor product of two 2D convolution kernels, one on the time and feature axes (red side of STS representation in Figure~\ref{fig:module_diagram}(a)), and one on the structure and feature axes (green side in Figure~\ref{fig:module_diagram}(a)), and we refer to the former as the {\em temporal kernel} and the latter as the {\em structural kernel}. The convolutional kernels of  the dual-stream CNN model are organized into a hierarchical structures, namely, the structural kernels are organized into different levels, (low, medium and high) to represent features of different scales; and the temporal kernels corresponding to the medium level structural kernels are put into categories of short, medium and long ranges. In addition, we apply a gating module for kernels of different sequential dependency ranges so the contributions of different feature ranges can be determined adaptively in a data-driven fashion.

We evaluate the performance of the dual-stream CNN model for STS classification on both synthetic data and practical data from skeleton based human activity recognition. The main contributions are summarized as follows: (1) Inspired by the dual-stream hypothesis in neural science, we propose a dual-stream CNN to learn various adaptive and flexible STS configurations, so as to boost the STS classification performance. (2) We design a novel flexible and scalable architecture for learning STS configurations, so we firmly believe that it will inspire other researchers to extend the presented ideas and to further advance the performance. (3) Experimental results on various benchmarks show that our method perform favorably against the state-of-the-art methods.

\section{Related Works}
\label{sec:bg}
A time series consists of a sequence of observations, one for each time step \cite{SantosK16}. Time series can be roughly categorized in terms of the observation dimensionality as the scalar (\eg,  \{($t_{1}$, 0.1), ($t_{2}$, 0.6), \ldots, ($t_{n}$, 0.3)\}) or the multivariate time series (\eg, \{($t_{1}$, <0.1, 0.3, 0.3>), ($t_{2}$, <0.2, 0.5, 0.1>), \ldots, ($t_{n}$, <0.8, 0.9, 0.6>)\}). The difference between STS and these two types lies not only in its higher observation dimensionality, but also in the key assumption regarding the dependencies among components within an observation, \ie, there are strong statistical dependencies among the components of each time step. For instance, in skeleton based activity recognition, the STS correspond to the joint locations, which can be represented as a tree structure, \eg, ($t_{i}$, <$d_{1}$, $d_{2} \rightarrow d_{1}$, $d_{3} \rightarrow d_{1}$, $d_{4} \rightarrow d_{2}$, \ldots>), where $\rightarrow$ pointing from the children to its parent reveals the tree structure, where components correspond to torso joints are placed on higher levels of the tree, and the spatial/temporal variation of a limb joint is constrained by its parent in the tree structure.

Classifying STS thus has to take into consideration such intertwining dependencies intra and inter time steps. Previous STS classification methods (see \cite{AggarwalX14} for a survey) focus on finding effective hand-crafted features for classification.  Most recent works have shifted to deep learning methods, and design neural networks for learning discriminative features automatically. Existing deep NN based methods for STS classification can be roughly divided into four categories.

\textbf{Feature Discriminability.}
Several works \cite{DuWW15,HuangWPG17,LiuSXW16,ZhangLXZXZ17}  designed a network structure follow the style of conventional works that took the inspiration of the STS spatial structural properties. Another set of works \cite{KeBASB17,LiuWHDK17,SongLXZL17,WengWY17} directly seeked statistically representative features from the STS data using attention mechanisms, gating, \etc. However, little effort has been made to integrate the hand-crafted and learned features for higher feature discriminability.

\textbf{Integration of Spatial and Temporal Information.}
Song \etal incorporated the LSTM with both the spatial and temporal attention mechanisms to adaptively select the discriminative dimensions and time steps for classification. Weng \etal \cite{WengWY17} adopted the bilinear classifiers to identify both key time steps and dimensions. Wang \etal~\cite{WangW17} formulated the STS as two sequences along the spatial and temporal axes, respectively, and used a two-stream RNN to model these two sequences. A common drawback of these work is that they assume independence between the structural and temporal information in the STS data.

\textbf{Sequential Orders.} STS instances satisfy the temporal causality, \ie, the observation at the current time step depends on those of the previous time steps. However, statistical dependency also exists in reverse of the time arrow. Thus, Du \etal \cite{DuWW15} used a bidirectional RNN structure to model STS. When the components of each observation at a time step can be represented by a tree structure, by traversing the tree structure bidirectionally using the depth-first search, the STS data can also be represented as a sequence spatially. Liu \etal \cite{LiuSXW16,LiuWHDK17} developed the spatiotemporal RNN for modeling such a spatial sequential order as an addition to the temporal sequential order.

\textbf{Sequential Dependency Ranges.} The range of temporal dependency is an important factor in modeling STS. This is commonly modeled using RNN with long-short term memory (LSTM) \cite{Graves2008}, in which the range is determined by the cooperation of the memory cell and several modulative gates. However, the capability of such an adaptive modeling would be overstretched given the high complexity of the STS data. To this end, several works explored to model the dependency ranges in a more controllable fashion. For example, Lee \etal \cite{LeeKKL17} incorporated the LSTM with multi-scale temporal sliding windows, so as to explicitly control the dependency ranges. Ke \etal \cite{KeBASB17} represented the STS data as 2D clips, and used the CNN to capture the long-range dependencies.

{\bf Temporal CNN}. RNN has been the dominant deep neural network structure for STS classification in the past few years.
However, a recent work \cite{GehringAGYD17} on machine translation suggests that a multi-layer convolutional neural network (CNN) is a more desirable option for modeling adaptive and flexible sequential order and sequential dependency in time series for two reasons. First, unlike the linear dependency model in RNN that the elements in a time series can only be processed in the forward or backward direction, temporal CNN can process multiple elements simultaneously, and the temporal dependency is no longer fixed universally but can be learned locally and adaptively during the training process. Second, the multi-layer temporal CNN creates hierarchical representations for time series in which nearby elements interact at lower layers while distant elements interact at higher layers; This provides a shorter path to capture long-range dependencies more effectively and efficiently than the chain structure modeled by RNN.

\section{Dual-stream Formulation with CNN}
\label{sec:prelim}
\label{sec:prelim:dualstream}
The temporal and structural dependency in the STS data are important patterns to be considered in STS classification. The temporal dependency is commonly modeled with a chain structure that goes in forward or backward direction, such as in dynamic Bayesian models or uni- or bi-directional RNN-LSTM. This is based on the \emph{sequential causality assumption} that the intrinsic temporal dependency of a sequence follows the time arrow or its reverse.

However, the sequential causality assumption does not always hold, as suggested by the indefinite causal order theory in quantum mechanics \cite{oreshkov2012quantum}, \ie, the causality order does not always obey a specific element permutation, but a mixture of multiple permutations. We refer to such a problem as the \emph{indefinite order problem}, which contradicts the sequential causality assumption made by RNN. In particular, for a RNN (unidirectional or bidirectional), long-range dependencies between two distant elements in a sequence might be affected by many other irrelevant elements on the long path through the chain. In addition, it is also hard for RNNs to accommodate such ``indefinite" permutations on the fly due to the indifferentiability of permuting operations.

Mitigating the long-range dependency modeling problem and indefinite order problem requires us to avoid using the chain structure but to create a STS classification model that is more flexible and adaptive, and to reduce the impacts caused by different element positions in a sequence. In this work, we model the sequential order of STS using the multi-layer CNN, which creates  hierarchical representations over the input STS in which the dependencies of nearby elements are modeled by lower layers while those of distant elements are modeled by higher layers. The replacement of RNN with CNN alleviates the two aforementioned problems, as the sequential dependency modeling is no longer strictly limited by a sequential order or a chain structure, but directly handled by the multi-scale receptive fields of CNN.  In the following, we describe the overall processing steps of the dual-stream CNN, starting with an augmented STS representation (Section \ref{sec:prelim:stsrep}) and then on the structure of the model itself (Section \ref{sec:2}).

\subsection{STS Representation}
\label{sec:prelim:stsrep}
We first augment the original STS data with empirical hand-crafted features as supplements to the original data before the feature learning process to form a rank-3 tensor $R_{d,t,f}$ and $R_{t,d,f}$ as follows. Given a STS $\{(t_{i}, <d_{1}, d_{2} \rightarrow d_{1}, d_{3} \rightarrow d_{1}, d_{4} \rightarrow d_{2}, \ldots>)\}_{i=1}^{n}$ with each dimension being a point $d_{j} = (x^{j}_{1}, x^{j}_{2}, \ldots, x^{j}_{l})$ in the $l$ dimensional space, we follow \cite{VeeriahZQ15,LiWCLL17} to extract four types of features for each dimension $d_{j}$ at each time step $t_{i}$, and concatenate them to form a feature vector $h^{t_{i}}_{d_{j}}$:
~\vspace{-1em}
\begin{enumerate} \itemsep -0.3em
  \item \textbf{Position.} $x^{j}_{1}, x^{j}_{2}, \ldots, x^{j}_{l}$ are concatenated to form a $l$ dimensional feature vector.
  \item \textbf{Angles.} Given multiple edges $\{e^{j}_{k}\}_{d_{k} \in \aleph_{j}}$ connecting $d_{j}$ and its neighboring dimensions $\aleph_{j}$ in the tree, we compute the normalized pairwise angles between these edges.
  \item \textbf{Offset.} Offsets of elements in $d_{j}$ between $t_{i}$ and $t_{i-1}$ are computed and concatenated to form a $l$ dimensional feature vector.
  \item \textbf{Distance.} We calculate the pairwise distance between $d_{j}$ and the mean position of all dimensions at $t_{i}$.
~\vspace{-1em}
\end{enumerate}
Then, we pad the extracted features to be equal in length, and concatenate the extracted features to form $R_{d,t,f}$ and $R_{t,d,f}$. The order of the STS dimensions in $R_{t,d,f}$ is determined by the bidirectional traversing algorithm \cite{LiuSXW16,LiuWHDK17} starting from $d_{1}$.

% !TEX root =  nips_2018.tex
\subsection{Model Structure}
\label{sec:2}
% configuration of the sequential order: e.g., hbrnn, ntu trust-gate, global context, etc.
% configuration of the contribution of spatial and temporal information: e.g.,
% configuration of the sequential dependency range: e.g, multi-scale (icip), etc.
% configuration of the features -> soft selective concatenation mechanism for soft feature selection: e.g., msra papers, differential RNN, global context, etc.

The temporal stream and structural stream in the dual-stream CNN model share similar structures. Given a STS in the tensor representation described previously, we process the 2D slices constituted by the time and feature elements of the STS tensor with the temporal stream CNN, and process the 2D slices constituted by the structure and feature elements of the STS tensor with the structural stream CNN.
%The computation of the convolution layer can be divided into two steps, \ie, the sliding of the convolution kernel and the addition of convolution results along the channel dimension. Convolution is parameterized by the convolution kernel, while the addition computation is non-parametric. Thus, we argue that the convolutional computation is more sensitive to modeling the statistical dependencies than the addition one.
For the temporal stream of our model, the convolutional computation on the time and feature axes captures the temporal dependencies, while the spatial dependencies are captured by the addition computation. The case is reversal for the structural stream. Thus, the desirable format of the input to the temporal stream should be $R_{t,d,f} = (h^{t_{1}}_{d_{j}}, h^{t_{2}}_{d_{j}}, \ldots, h^{t_{n}}_{d_{j}})_{j=1}^{m}$, where $h^{t_{i}}_{d_{j}}$ represents the feature vector of the $j$th dimension at the $i$th time step. Similarly, the input to the structural stream is $R_{d,t,f} = (h^{t_{i}}_{d_{1}}, h^{t_{i}}_{d_{2}}, \ldots, h^{t_{i}}_{d_{m}})_{i=1}^{n}$ which is a transpose of $R_{t,d,f}$.

\begin{wrapfigure}{r}{0.52\textwidth}
%\vspace{-1cm}
\centering
\includegraphics[width=0.52\textwidth]{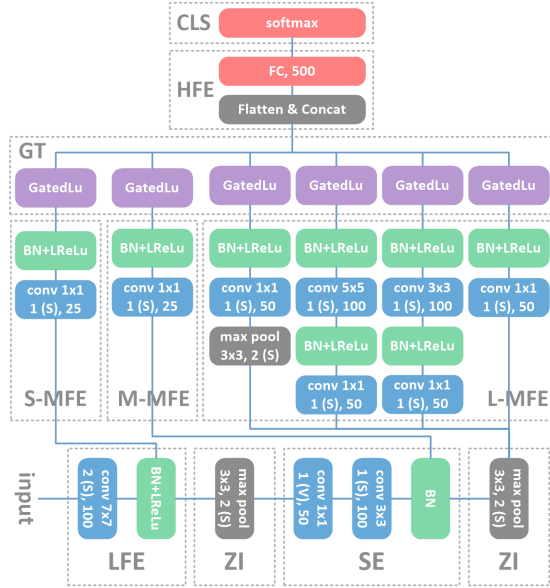}
~\vspace{-1.5em}
\caption{\small \em The network structure of a single stream of our model with ten building blocks.}
\label{fig:network_structure}
~\vspace{-1.5em}
\end{wrapfigure}

The preprocessed STS features are fed into the dual-stream CNN for feature learning. The adaptive selection of sequential dependency range is an important factor in STS classification. In our work, we fuse the adaptive learning of the sequential dependency range into the feature learning process, with the purpose of making these two phases boost each other.
There are ten building blocks within our dual-stream framework as shown in Figure~\ref{fig:network_structure}, including low/medium/high-level feature extractor (L/M/HFE), two zoom-in modules (ZI), a shared encoder (SE), a gating module (GT) and a classification module (CLS). The MFE is composed of three sub-blocks, \ie, short/medium/long-range MFE. The explanation of these blocks is given below.

As shown in Figure~\ref{fig:network_structure}, we decompose the feature learning process into three stages, \ie, low/medium/high-level feature extractors (L/M/HFE). The low-level features keep more details of the original input STS data, while the high-level features are more conceptual which are used for classification directly. The medium-level features bridge the low-level and the high-level features, so it determines the reliability and the meaningfulness of the high-level features. In the following, we discuss these three feature extractors in detail.

\textbf{LFE}. This module consists of a convolutional layer with kernel size 7 and the batch normalization (BN) followed by a leaky relu (LReLu). The reason why we chose a large kernel size for LFE is due to the high similarity of consecutive elements in a sequence which may contains much redundancy. Note that we use LReLu as the nonlinearity in our neural network which does not gate the negative values in our STS representation.

\textbf{MFE.} Followed by the LFE is the MFE, which MFE plays the key role in feature learning, and accordingly the design of the MFE is the most complicated among these three feature extractors. MFE is decomposed into three sub-stages with each focusing on a specific sequential dependency range corresponding to the short/medium/long-range MFE (denoted as S/M/L-MFE). The MFEs for different dependency ranges are connected by zoom-in (ZI) modules and shared encoders (SE). ZI is implemented as a max pooling layer, and SE is an encoder block posed at shared by the M-MFE and L-MFE. Since the space covered by L-MFE is larger than that of short/medium-range MFEs, we further split L-MFE into four finer scales similar to the inception module of the work in \cite{SzegedyLJSRAEVR15}.

\textbf{GT}. We place a gating module (GT) as the backend of S/M/L-MFE which adaptively determines the contribution of each sequential dependency range to the high-level features. We implement the gating module as the gated linear unit (GatedLu) \cite{DauphinFAG17} over the output of the convolution $Y = [A~B] \in \mathbb{R}^{w,h,2c}$, $v([A~B]) = A \otimes \sigma(B)$, where $A,B \in \mathbb{R}^{w,h,c}$ are the inputs to the non-linearity, $\otimes$ is the point-wise multiplication and the output $v([A~B]) \in \mathbb{R}^{w,h,c}$ is half size the size of $Y$. The gates $\sigma(B)$ (implemented as a sigmoid function) control which inputs $A$ of the current medium-level features are helpful for learning the high-level features.

\textbf{HFE}. The HFE module is formed with a fully connected layer (FC) which takes the flattened and concatenated medium-level features output by S/M/L-MFE, and outputs a 500 dimensional high-level feature vector for each STS.

\textbf{Classification}.
The extracted high-level features are fed into a softmax layer to obtain the probability distributions, which are used to compute the negative log-likelihood loss, $L = -\sum_{i} y^{\prime}_{i} \log (y_{i})$, where $y_{i}$ is the probability distribution output of an input STS by the softmax layer, and $y^{\prime}$ is its corresponding ground truth one-hot vector representation.

% !TEX root =  nips_2018.tex
\section{Experiments}
\label{sec:exp}

\begin{wraptable}{R}{0.52\textwidth}
\vspace{-1cm}
	 \begin{minipage}{0.52\textwidth}
\begin{algorithm}[H]
\small
\caption{Synthesize STS data}
\label{alg:Alg1}
\begin{algorithmic}
\REQUIRE Number of classes $N$, number of instances per class $M$, temporal length $T$, rate of change $\Delta$
\ENSURE $M$ STS instances for each of $N$ classes
\STATE $i \longleftarrow 0$
\FOR{$i < N$}
    \STATE Sample the change range of the distance $[\theta_{l}, \theta_{u}]$ with $\theta_{l} \in [0,4]$ and $\theta_{u} \in [\theta_{l},\theta_{l}*\Delta]$, and the change range of the angle of $i$th class $[\phi_{l}, \phi_{u}]$ with $\phi_{l} \in [0,0.2]$ and $\phi_{u} \in [\phi_{l},\phi_{l}*\Delta]$
    \STATE $j \longleftarrow 0$
    \FOR{$j < M$}
        \STATE Sample the $x$-$y$ coord of a virtual root dimension from $\mathcal{N}(0,1)$ for whole sequence
        \STATE Recursively synthesize the dimensions by sampling the angle from $\mathcal{N}(0,10)$ and the distance from $\mathcal{N}(5,1)$ for the first time step
        \STATE Sample the distance and angle change per time step for each dimension from $[\theta_{l}, \theta_{u}]$ and $[\phi_{l}, \phi_{u}]$
        \STATE $k \longleftarrow 1$
        \FOR{$k < T$}
            \STATE Move the dimensions according to the sampled distance and angle changes
        \ENDFOR
    \ENDFOR
\ENDFOR
\end{algorithmic}
\end{algorithm}
\end{minipage}
\vspace{-1.2cm}
\end{wraptable}

We evaluate the performance of the dual-stream CNN model on artificially generated STS data, and real STS data from the skeleton based activity recognition problem.

\subsection{Experimental Settings}
Our method is implemented using Python and Google TensorFlow \cite{AbadiBCCDDDGIIK16}, and all experiments are conducted on four machines on each of which an NVIDIA TITAN X GPU with 12GB onboard memory is installed. The overall objective function is minimized using back-propagation implemented with the ADAM algorithm \cite{KingmaB14}. We train the network using mini-batch gradient descent, and set learning rate, momentum and decay rate as $1 \times 10^{-3}$, $0.9$, $0.999$. As usual, we scale the input to be equal in temporal length, so as to enable the mini-batch processing. To evaluate the STS classification performance of our method, we conduct experiments on both a general domain with the synthetic STS data ($\S$~\ref{sec:exp:synthetic}) and skeleton based human activity recognition on several widely used benchmarks ($\S$~\ref{sec:exp:sar}).

\subsection{Experiments on Synthetic STS}
\label{sec:exp:synthetic}
\begin{figure*}
\centering
\includegraphics[width=1\textwidth]{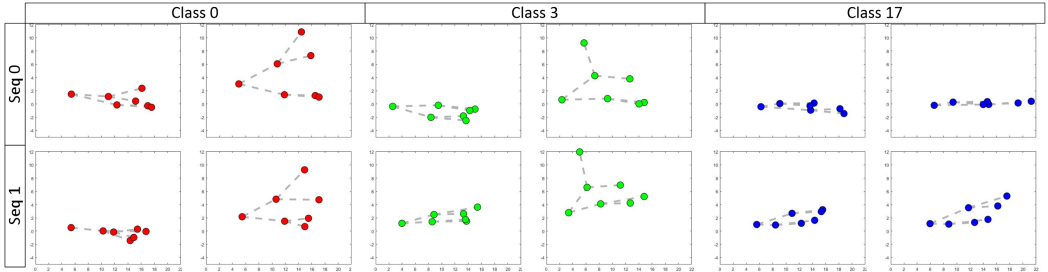}
~\vspace{-1em}
\caption{\small \em The illustration of the synthetic STS data. We show two frames of three classes.}
\label{fig:synthetic_data}
~\vspace{-1.5em}
\end{figure*}

\textbf{Data Generation.}
We synthesized $6,000$ STS data, which correspond to $60$ classes with $100$ instances within each class. For each STS instance, at each time step, we sample seven points on the $x$-$y$ plane as an observation for that time step, which are organized as the node of a three-level binary tree. Specifically, we represent each point using its polar coordinate relative to its parent, \ie, the distance away from its parent and the angle between the $x$ axis and the vector pointing from the current point to its parent. Note that we cannot represent the root dimension using with the polar coordinate, since it does not have the parent. Thus, in order to make the root dimension movable, we create a virtual root dimension as its parent, and the position of the virtual root remains the same for the whole sequence. We illustrate three classes of the synthetic data in Figure~\ref{fig:synthetic_data}. The inter-class variations of the synthetic data are in the temporal changes of the ranges of distances and angles of the child node. The data generation process is given in Algorithm~\ref{alg:Alg1}.

\textbf{Baselines.} We implement nine baseline classifiers including the random forest (RF), k-nearest neighbor (KNN), decision tree (DT), gaussian naive bayes (GNB), quadratic discriminant analysis (QDA), multi-layer perception (MLP), support vector machine (SVM), one-layer LSTM, and three-layer CNN.

\textbf{Analysis.} We use 70$\%$ of the data for training, and the rest for test. Table~\ref{Tab1} shows the comparison between the baselines and our method on the synthetic STS data. Our method outperforms all baselines by a large margin. We observe that several evaluated methods (\ie, SVM, QDA, LSTM, CNN) can achieve perfect training accuracy, \ie, nearly $1.0$. However, only our method achieves the test accuracy greater than $0.9$. This shows the good generalization of our method despite the complicated architecture, which is owing to the adaptability achieved by the BN and GatedLU modules in our model. In addition, there are almost no hyper-parameters in our method that need tuning, which reflects our method's robustness.

\subsection{Experiments on Skeleton based Activity Recognition}
\label{sec:exp:sar}
For a real life application, we chose to apply the dual-stream CNN model to human action recognition, which is a STS classification problem. Specifically, the skeleton based activity recognition problem is formulated as categorizing the sequences composed of human skeletons into pre-defined classes. We tested our method on four challenging datasets: MSR Action3D \cite{LiZL10}, CharLearn Italian \cite{EscaleraGBRLGAE13}, 3D-SAR-140 \cite{LiWCLL17} and NTU RGB+D \cite{ShahroudyLNW16}.

\begin{wraptable}{r}{0.2\textwidth}
\vspace{-0.7cm}
\caption{\small \em Performance of various methods on synthetic STS.}
\centering
\footnotesize
\label{Tab1}
\begin{tabular}[t]{|p{1cm}|c|}\hline
{Methods} &{Accur.}\\ \hline\hline
{RF} &0.220 \\ \hline
{KNN} &0.281 \\ \hline
{DT} &0.286 \\ \hline
{GNB} &0.432 \\ \hline
{QDA} &0.474 \\ \hline
{MLP} &0.653 \\ \hline
{LSTM} &0.769 \\ \hline
{CNN} &0.789 \\ \hline
{SVM} &0.842 \\ \hline
{Ours} &{\textbf{0.922}} \\ \hline
\end{tabular}
\vspace{-0.7cm}
\end{wraptable}

\textbf{MSR Action3D.} This dataset consists of $20$ actions performed by $10$ subjects in an unconstrained way for two or three times, $557$ valid samples with $22,077$ frames. All sequences are captured in $15$ FPS, and each frame in a sequence contains $20$ skeleton joints. We follow the experimental protocol presented in \cite{WangLWY12} on this dataset, which is the most challenging protocol for this dataset. Half of actor subjects are used for training and the rest are used for test. Note that the average number of training samples per class is nearly 14, which is quite limited for training deep neural networks, and poses great potential risks on the overfitting issue. The comparison on MSR Action3D in Table~\ref{Tab2} demonstrates the good generalization capability of our method under a more challenging circumstance compared to the synthetic dataset.

\textbf{CharLearn Italian.} This dataset captures $20$ Italian cultural signs, and contains $393$ labeled sequences with a total of $7,754$ gesture instances. We follow the experimental protocol in \cite{WuS14}: $350$ sequences for training and the rest $43$ sequences for testing (each sequence contains $20$ unique gestures). The recognition of sign languages require the fine-grained recognition ability of the evaluated methods, and always desire the careful feature representation design. As shown by the comparison on CharLearn Italian in Table~\ref{Tab2}, our method constantly outperforms the evaluated methods without designing any special features for his dataset.

\textbf{3D-SAR-140.} This dataset contains $140$ diverse action classes by aggregating all distinct classes from $10$ existing datasets: CMU Mocap~\cite{CMUMOCAP}, ChaLearn Italian~\cite{EscaleraGBRLGAE13}, MSRC-12 Gesture~\cite{FothergillMKN12} (12), MSR Action3D~\cite{LiZL10}, HDM05~\cite{cg-2007-2}, Kintense~\cite{NirjonGTZSYRBPS14}, Berkeley MHAD~\cite{OfliCKVB13}, MSR Daily Activity 3D~\cite{WangLWY12}, UTKinect-Action~\cite{XiaCA12}, and ORGBD~\cite{YuLY14}. 3D-SAR-140 is a challenging benchmark due to two factors: (1) a large variety of movements and dynamics in various contexts are included, where fine-grained recognition is required; (2) sequence length for individual actions varies significantly (ranging from $5$ to $800$ frames) within or across classes, which poses the challenges on the adaptive configuration of the sequential dependency range. Notably, URNN-2L-T \cite{LiWCLL17} is designed to recognize fine-grained actions, and handle the similarity and dissimilarity among large-scale diverse classes. However, our method still performs slightly better than URNN-2L-T. The good performance also demonstrates our method's effectiveness in adaptively configuring the sequential dependency ranges.

\textbf{NTU RGB+D.} This dataset \cite{ShahroudyLNW16} contains more than $56,000$ samples, and includes $60$ classes. To our knowledge, this is the largest skeleton based activity recognition dataset. There are two standard evaluation protocols: (1) cross-subject: $20$ subjects are used for training, and the remaining $20$ subjects are for testing; (2) cross-view: two view-points are used for training, and one is for testing. The large amount of variations in subjects and views make this dataset challenging, so some methods, \eg, \cite{KeBASB17,WangW17}, employ the data augmentation mechanism or learn the 3D transformation model \cite{ZhangLXZXZ17} to improve the generalization. This is the key reason why \cite{ZhangLXZXZ17} performs slightly better than our method under the cross-view setting. Note that our method can still achieve the favorable performance, though we do not augment or transform the data. It is clear that our method performs better than most state-of-the-art methods on NTU RGB+D that are based on the explicit spatiotemporal modeling \cite{LiuWHDK17,SongLXZL17,WangW17} or RNN/LSTM \cite{LeeKKL17,LiWCLL17,LiuSXW16,LiuWHDK17,ShahroudyLNW16,SongLXZL17,WangW17}. This reveals the benefits brought by the dual-stream network and the temporal CNN architecture in modeling the interweaving spatiotemporal modeling and adaptive configuration of sequential orders.

\begin{table}
~\vspace{-1em}
\caption{\small \em Classification accuracy on MSR Action3D \cite{LiZL10}, CharLearn Italian \cite{EscaleraGBRLGAE13}, 3D-SAR-140 \cite{LiWCLL17} and NTU RGB+D \cite{ShahroudyLNW16}. The ``CS" and ``CV" represent the cross-subject and cross-view settings, respectively.}
~\vspace{-1em}
\footnotesize
\begin{tabular}[t]{|p{3.1cm}|c|c|c|}\hline
{Methods} &{\cite{LiZL10}} &{\cite{EscaleraGBRLGAE13}} &{\cite{LiWCLL17}}\\ \hline\hline
{RR \cite{VemulapalliC16}} &0.891 &0.438 &0.723\\ \hline
{HBRNN-L \cite{DuWW15}} &0.897 &0.559 &0.604 \\ \hline
{CHARM \cite{LiWCL15}} &0.747 &0.476 &0.618 \\ \hline
{DBN-HMM \cite{WuS14}} &0.735 &0.628 &0.601 \\ \hline
{Lie-group \cite{VemulapalliAC14}} &0.866 &0.401 &0.745 \\ \hline
{HOD \cite{GowayyedTHE13}} &0.844 &0.539 &0.657 \\ \hline
{MP \cite{ZanfirLS13}} &0.909 &0.452 &0.203 \\ \hline
{SSS \cite{ZhaoLPZS13}} &0.560 &0.413 &0.253 \\ \hline
{HBRNN-L-T \cite{LiWCLL17}} &0.915 &0.673 &0.756 \\ \hline
{URNN-2L-T \cite{LiWCLL17}} &0.931 &0.753 &0.892 \\ \hline\hline
{Ours $\ominus$ gating module} &0.947 &0.766 &0.864 \\ \hline
{Ours $\ominus$ structural stream} &0.848 &0.677 &0.814 \\ \hline
{Ours $\ominus$ temporal stream} &0.934 &0.729 &0.889 \\ \hline
{Ours} &\textbf{0.963} &\textbf{0.772} &\textbf{0.896} \\ \hline
\end{tabular}
\hfill
\begin{tabular}[t]{|p{3.1cm}|c|c|}\hline
{Methods} &{CS \cite{ShahroudyLNW16}} &{CV \cite{ShahroudyLNW16}}\\ \hline\hline
{ST-LSTM \cite{LiuSXW16}} &0.692 &0.777 \\ \hline
{LieNet \cite{HuangWPG17}} &0.614 &0.670 \\ \hline
{Two-strem RNN \cite{WangW17}} &0.713 &0.795 \\ \hline
{STA-LSTM \cite{SongLXZL17}} &0.734 &0.812 \\ \hline
{GCA-LSTM \cite{LiuWHDK17}} &0.744 &0.828 \\ \hline
{Ensemble \cite{LeeKKL17}} &0.746 &0.813 \\ \hline
{URNN-2L-T \cite{LiWCLL17}} &0.746 &0.832 \\ \hline
{Bi-modal \cite{RahmaniB17}} &0.752 &0.831 \\ \hline
{VA-LSTM \cite{ZhangLXZXZ17}} &0.794 &\textbf{0.876} \\ \hline
{MTLN \cite{KeBASB17}} &0.796 &0.848\\ \hline\hline
{Ours $\ominus$ gating module} &0.734 &0.809 \\ \hline
{Ours $\ominus$ structural stream} &0.722 &0.791 \\ \hline
{Ours $\ominus$ temporal stream} &0.779 &0.850 \\ \hline
{Ours} &\textbf{0.811} &0.872 \\ \hline
\end{tabular}
\label{Tab2}
%~\vspace{-1em}
\end{table}

\textbf{Ablation Study.} We evaluate three components in our method, \ie, the gating module, structural stream and temporal stream. We disable these components one by one and evaluate them on skeleton based activity recognition datasets. Table~\ref{Tab2} shows the comparison results, and it clearly shows that each of these three components is beneficial for the generalization. Generally speaking, the descending order of their influences is structural stream, gating module and temporal stream. An interesting phenomenon is that the structural stream seems to have more important effect than the temporal stream in the final classification performance, which accords with the theory of dual-stream hypothesis in neural science. This strengthens the reasonability of our inspiration drawn from the dual-stream hypothesis, and further shows the necessity of interweaving spatiotemporal modeling for STS classification.

% !TEX root =  nips_2018.tex
\section{Conclusion}

The structured time series (STS) classification problem requires the modeling of interweaved spatiotemporal dependency. most previous STS classification methods model the spatial and temporal dependencies independently. In this work, inspired by the dual-stream hypothesis in neural science, we propose a novel dual-stream framework for modeling the interweaved spatiotemporal dependency, and develop a convolutional neural network within this framework that aims to achieve high adaptability and flexibility in STS configurations from various diagonals, \ie, sequential order, dependency range and features. The proposed architecture is highly modularized and scalable, making it easy to be adapted to specific tasks. The effectiveness of our model is demonstrated through experiments on synthetic data as well as benchmark datasets for skeleton based activity recognition.

For future works, we plan to further extend the current work in the following aspects. First, we will explore more real-life applications of STS classification, for instance, financial time series. Second, in order to reduce the human labors in labeling the data, we will investigate the semi-supervised learning for STS classification based the mixture of labeled and unlabeled data. This will reduce the requirement on labeled STS data and improve the overall performance of the algorithm.

\newpage
\bibliographystyle{ieee}
\bibliography{arxiv_dualCNN}

\end{document}